\documentclass{article}
\usepackage{caption}
\usepackage{amsmath}
\usepackage{cite}
\usepackage{graphicx}
\usepackage{float}
\usepackage{subfigure}
\usepackage{booktabs}
\usepackage[table,xcdraw]{xcolor}
\usepackage{float}
\usepackage{marvosym}
\usepackage[T1]{fontenc}
\usepackage{aecompl}
\usepackage{spconf,amsmath,epsfig}

\let\OLDthebibliography\thebibliography
\renewcommand\thebibliography[1]{
  \OLDthebibliography{#1}
  \setlength{\parskip}{0pt}
  \setlength{\itemsep}{0pt plus 0.3ex}
}
\begin{document}\sloppy

\def\x{{\mathbf x}}
\def\L{{\cal L}}

\title{Attention Mechanism for Contrastive Learning in GAN-based \\Image-to-Image Translation}
%
\name{Hanzhen Zhang, Liguo Zhou*, Ruining Wang, Alois Knoll}
\address{Chair of Robotics, Artificial Intelligence and Real-time Systems, Technical University of Munich\\liguo.zhou@tum.de}

\maketitle

\begin{abstract}
Using real road testing to optimize autonomous driving algorithms is time-consuming and capital-intensive. To solve this problem, we propose a GAN-based model that is capable of generating high-quality images across different domains. We further leverage Contrastive Learning to train the model in a self-supervised way using image data acquired in the real world using real sensors and simulated images from 3D games. In this paper, we also apply an Attention Mechanism module to emphasize features that contain more information about the source domain according to their measurement of significance. Finally, the generated images are used as datasets to train neural networks to perform a variety of downstream tasks to verify that the approach can fill in the gaps between the virtual and real worlds.
\end{abstract}
\section{Introduction}
\label{sec:intro}

With the development of deep learning, rapid progress has been made in autonomous driving technology. However, training the deep learning model needs a massive amount of labeled real road data. Autonomous driving vehicles also require to go through a lot of road testing before they can actually be used in transportation. Using real road testing to optimize autonomous driving algorithms is time-consuming and capital-intensive, and testing on open roads has safety implications and regulatory constraints. In addition, extreme weather conditions and complex traffic scenarios are difficult to be reproduced in reality. Therefore, it makes sense to convert simulator image data into photo-realistic image data, which can save time and capital costs and help the research and implementation of autonomous driving algorithms. 

\begin{figure*}[h]
    \centering
    \subfigure[High-poly 3D Models]{
    \label{Fig.1.1}
    \includegraphics[height=2.7cm]{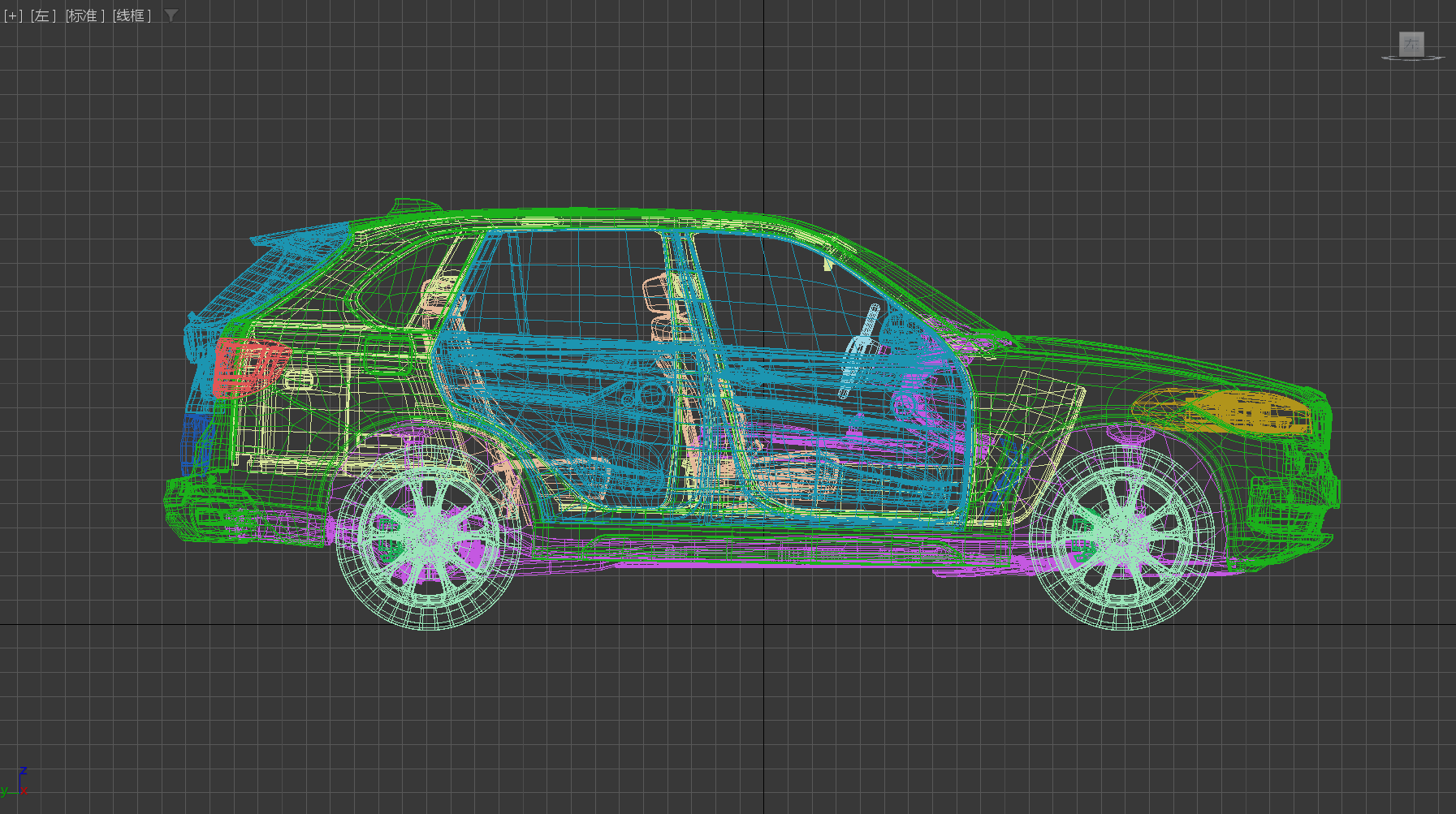}}
    \subfigure[High Definition Rendering]{
    \label{Fig.1.2}
    \includegraphics[height=2.7cm]{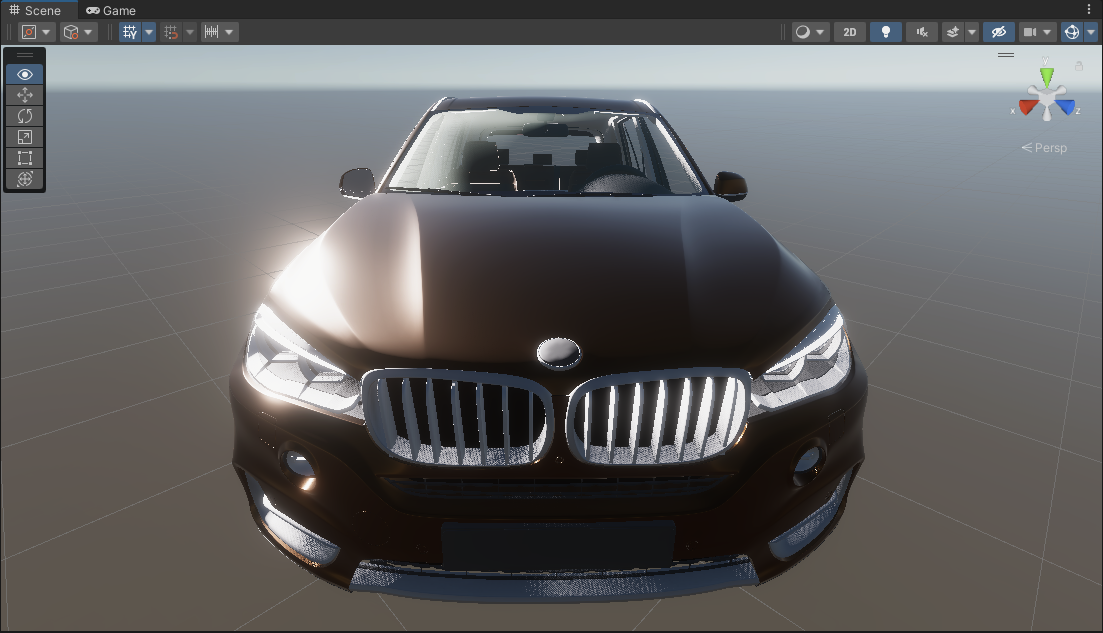}}
    \subfigure[GAN-based Post-processing]{
    \label{Fig.1.3}
    \includegraphics[height=2.7cm]{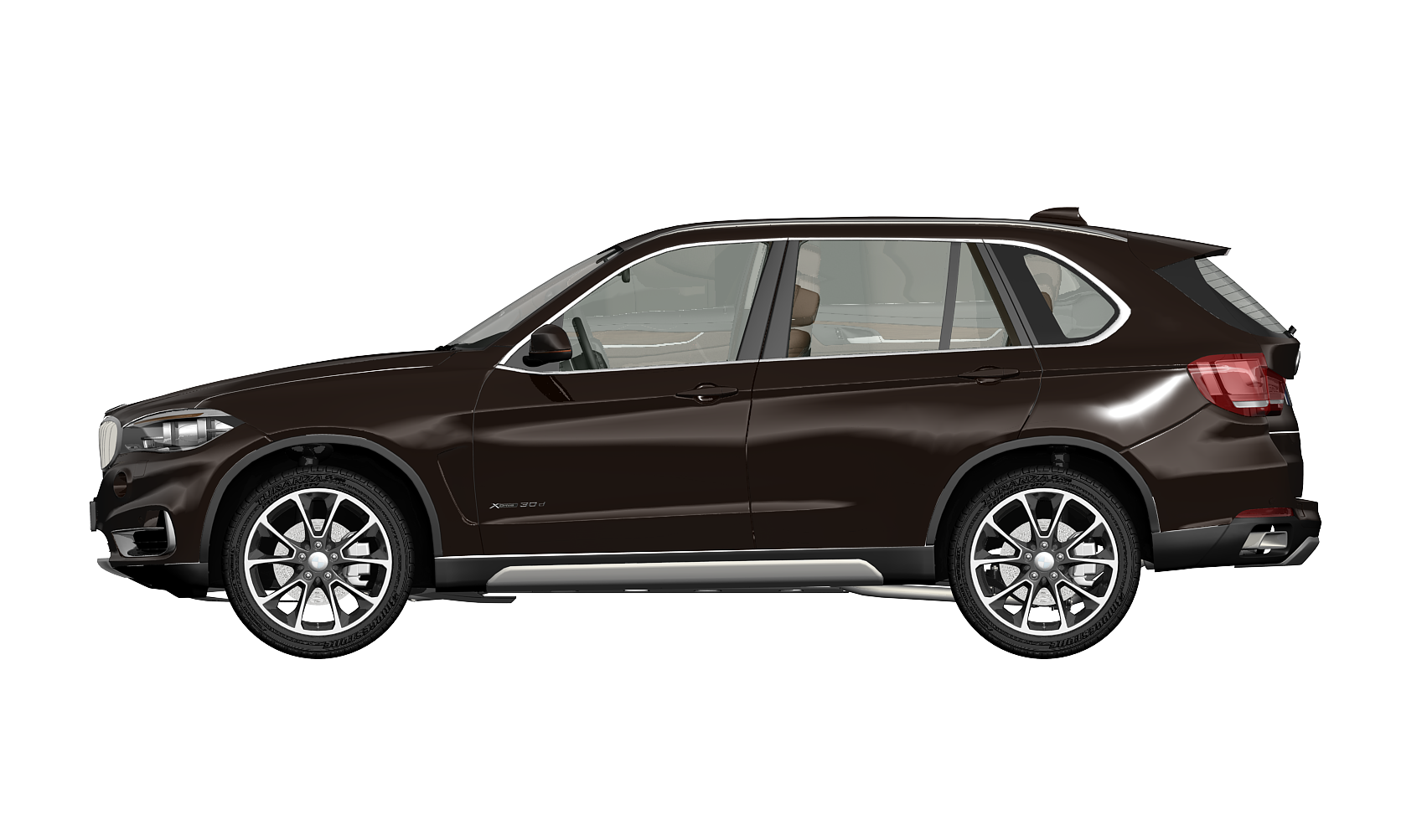}}
    \caption{Three Steps towards Photo-realistic Virtualization of Road Object}
    \label{figure:bmw}
\end{figure*}

Figure \ref{figure:bmw} shows the three steps toward Photo-realistic Virtualization of road objects. In this paper, we focus on solving the third step towards the photo-realistic virtualization of road objects, namely the GAN-based post-processing algorithms. Specifically, we need to design a suitable generator that can learn the joint distribution of the two domains (real and simulated world) and find transformations between them. Considering the difficulty of acquiring paired images, i.e. generated labels, we need to train the generator with unpaired images for self-supervised learning. Moreover, in order to maximize the mutual information between the source and the translated images across different domains, we also need to solve the problem that Generators need to be able to identify important information and features.

The main contributions of this paper can be summarized as follows: 

\begin{itemize}
    \item We introduce the main modules and Attention Mechanism module into our Attention-based CUT\_GAN. Then we evaluate the performance of the designed model on several datasets to check its validity.
    \item We compare several attention mechanisms through ablation experiments to verify the improvement of different attention mechanisms on the performance of generators. Furthermore, the effectiveness of the attention mechanism module is verified by comparing the designed model with other generative algorithms.
\end{itemize}

\section{Related works}
In this section, we make a comprehensive review of the development of the methods used in this paper, including Generative Adversarial Networks, Image-to-Image Generators, and self-supervised methods.

\subsection{Generative Adversarial Networks}
Since Ian GoodFellow proposed the GAN model \cite{goodfellow2020generative} in 2014, generative adversarial networks have rapidly become the hottest generative model. CGAN (conditional GAN)\cite{perarnau2016edit} adds constraints to the original GANs so that the network can generate samples in a given direction. The proposal of DCGAN (deep convolutional GAN)\cite{radford2015dcgan} has great significance to the development of GAN, which combines convolutional neural networks (CNN) and GAN. It combines CNN and GAN to ensure the quality and diversity of the generated images. LSGAN (least squares GAN)\cite{mao2017lsgan}, also known as least squares generative adversarial network, replaces the cross-entropy loss function of a traditional GAN with a least squares loss function, which effectively improves the problems of low quality and unstable training of the original GANs. 
CycleGAN\cite{zhu2017unpaired} allows images from two domains to be transformed into each other and does not require paired images as training data.
BigGAN\cite{brock2018biggan} gets its performance boost by scaling up the model, and is so far the best GAN model for generating image quality.

\subsection{Image-to-image Translation}
The image-to-image translation methods can be classified into two classes: Supervised methods and Unsupervised methods.

Supervised methods such as Pix2pix\cite{isola2017image} can translate images from one domain to another e.g. from night to day, from sketch to real picture and so on, but images must be paired. 

In unsupervised methods, Cycle GAN\cite{zhu2017unpaired} first translates an image from a source domain A to a target domain B in the absence of paired examples, then couples it with an inverse mapping: domain B to domain A and uses cycle consistency loss to let generated A image the same as the original image as much as possible. In ACGAN\cite{odena2017conditional}, the discriminator not only discriminates whether the input image is from the distribution of the generated data or from the distribution of the real data, but also predicts the class of the input image.

\subsection{Self-supervised Learning}
Self-supervised learning differs from unsupervised learning mainly in that unsupervised learning focuses on detecting specific data patterns, such as clustering, community discovery, or anomaly detection, while self-supervised learning aims at recovering and remaining within the supervised learning paradigm.

According to Jie Tang et al.\cite{liu2021ssl}, self-supervised learning can be divided into three general categories: Generative, Contrastive, and Generative-Contrastive. ContraGAN\cite{kang2020contragan}, which uses Contrastive Learning, tries to draw close to image representations that have the same category label, when considering drawing close to the same transformed image.

\section{Attention$-$based CUT\_GAN}

In this section, we introduce the structure of our proposed Image-to-Image Translation Network called Attention-based CUT\_GAN. Sections 3.2 and 3.3 introduce the two parts of Generative adversarial networks: Generator and Discriminator. Section 3.4 describes Patchwise Contrastive Learning and other parts of the loss function.

\subsection{Network Overview}
\begin{figure}[h]
    \centering
    \includegraphics[width=0.4\textwidth]{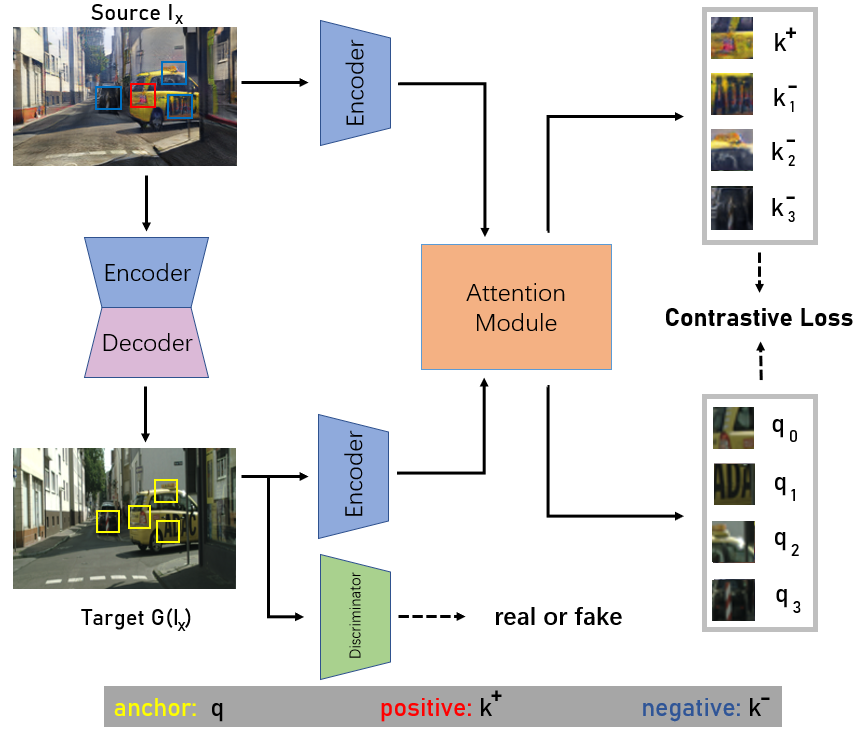}
    \caption{Overview of the proposed Image-to-Image Translation Network}
    \label{fig:overflow}
\end{figure}

The overflow of our proposed Image-to-Image Translation Network is depicted in figure \ref{fig:overflow}. We wish to translate images from input domain $\chi \subset R^{H\times W\times 3 }$ to appear like an image from the output domain $Y \subset R^{H\times W\times 3 }$. We are given a dataset of unpaired instances $X = $ \{$x\in \chi$\}, $Y = $\{$y \in Y$\}. Our method only requires learning the mapping in one direction and avoids using inverse auxiliary generators and discriminators. This can largely simplify the training procedure and reduce training time.

Given an image $I_{x}$, the target image $G(I_{x})$ can be obtained by the generator, which consists of an Encoder and Decoder. Meanwhile, the discriminator will determine whether the generated image is fake or real. Based on the original GAN foundation, our proposed network extracts the Encoder in the generator and uses it to extract the feature maps of the source image $I_{x}$ and the target image $G(I_{x})$ respectively. At this point, due to the dimensionality reduction operation of convolution, the feature map can be seen as dividing the image into patches of the same size as the perceptual field.

The original CUT\_GAN randomly samples patches in the feature map to compute the Contrastive loss. However, not all patches have rich domain information, so random sampling causes the GAN network to not maximize the information learned between two domains. To address this problem, we propose an additional Attention Module to intentionally select more significant patches for computing the Contrastive loss. The attention mechanism module calculates the significance of each patch, and by ranking them we select patches that are more important, i.e. contain more information about the domains.

Finally, all these sampled patches are used to compute the self-supervised contrastive loss. This loss term encourages two elements (corresponding patches) to map to a similar point in a learned feature space, relative to other elements (patches) in the dataset, referred to as negatives. The output can as results take on the appearance of the target domain while retaining the structure, or content, of the specific input.

\subsection{Generator}
The generator G  is designed to map the latent space vector (z) to data space. Since our data are images, converting z to data space means ultimately creating an RGB image with the same size as the training images (i.e. 3$\times$64$\times$64).

Empirically, the depth of the network is critical to the performance of the model. When the number of layers is increased, the network can perform more complex feature pattern extraction, so theoretically better results can be achieved when the model is deeper. However, experiments have revealed a degradation problem in deep networks: as the depth of the network increases, the accuracy of the network saturates and even decreases.

To address this problem, we propose a ResNet-based generator G\cite{johnson2016resnet-based} consists of two down-sampling blocks, nine intermediate blocks, two up-sampling blocks and two convolution layers. We apply Instance Normalization (IN)\cite{ulyanov2016instance}\cite{ioffe2015batch} and ReLU\cite{maas2013relu} in the generator, except for the output layer, which uses Tanh as the activation function. The network before the sixth residual block is regarded as the encoder E and the rest is the decoder. We adopt the multi-layer feature extraction in CUT\_GAN\cite{park2020cut}, which takes the features from five layers. including the input image, the first and second down-sampling blocks, and the first and fifth residual blocks. 

\subsection{Discriminator}
The discriminator D is a binary classification network that takes an image as input and outputs a scalar probability that the input image is real (as opposed to fake).

\begin{figure}
    \centering
    \includegraphics[width=0.4\textwidth]{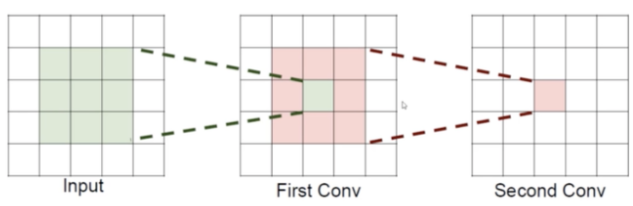}
    \caption{Perceptual field}
    \label{fig:perceptual}
\end{figure}

The discriminator in our proposed network uses a design called PatchGAN\cite{ma2019patchgan}. In general, the discriminator of the original GAN is designed to output only one boolean value (True or False), which is an evaluation of the whole image generated by the generator. In contrast, PatchGAN is designed to be fully convolutional, where the image is mapped using convolution into an $N \times N$ matrix, which is equivalent to the final evaluation value in the original GAN. Each point (true or false) in the $N \times N$ matrix represents a small region (in the sense of patch) of the original image, which is the application of the perceptual field, which is shown in Figure \ref{fig:perceptual}. Instead of measuring the whole image with a single value, the whole image is now evaluated using an $N \times N$ matrix, which obviously allows more areas to be focused on, which is the advantage of PatchGAN.

\subsection{Patchwise Contrastive Learning}
Image translation needs to ensure consistency of content. More precisely, important elements of the image such as pedestrians, cars, traffic lights, and even road patterns should be preserved as much as possible. Contrastive learning can then solve this problem. Specifically, a patch is selected from the real image as query; a number of patches are selected from the simulated image, and the patches with the same position as query are used as positive samples, while the others are used as negative samples.

Inspired by \cite{gutmann2010nce}\cite{chen2016infogan}\cite{park2020cut}, we use PatchNCE\cite{park2020cut} as contrastive loss. NCE, which is called "Noise Contrastive Estimation", estimates model parameters and normalization constants by maximizing the same objective function. NCE converts the problem into a binary classification problem, where the classifier is able to bifurcate the data samples and the noise samples. In addition, the negative samples for contrastive learning are selected patches within a single input image, rather than other images from the dataset.

\begin{figure}[h]
    \centering
    \includegraphics[width=0.45\textwidth]{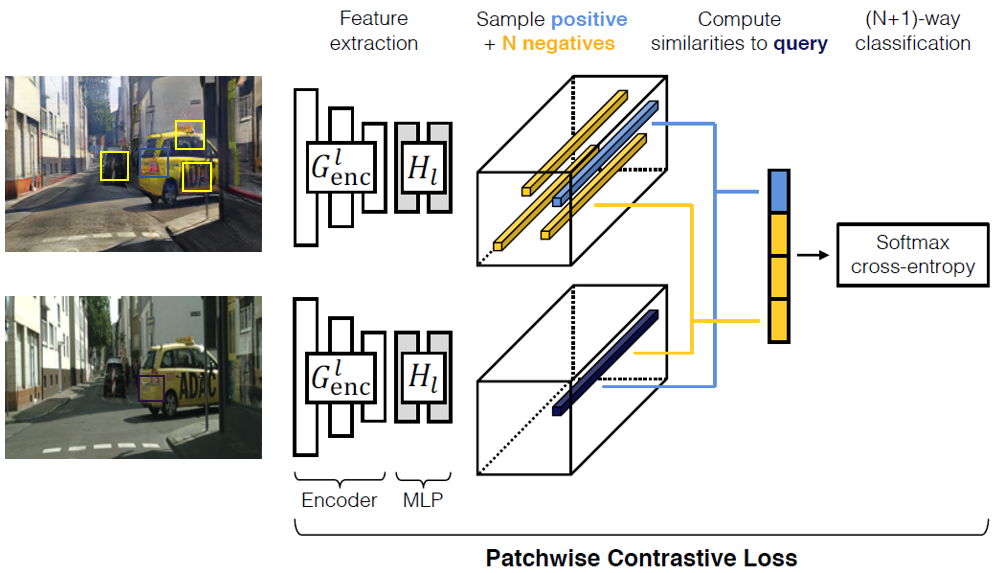}
    \caption{Patchwise Contrastive Loss}
    \label{fig:contrastive}
\end{figure}

Figure \ref{fig:contrastive} depicts the structure of Patchwise Contrastive Loss. As mentioned before, a generator is an encoder-decoder architecture, which passes the real and simulated images through this encoder, and selects the feature maps output from the L layers of the encoder to calculate the contrastive loss in the feature space instead of the original image, as shown in

\begin{equation}
    L _{PatchNCE}(G, H, X) = E_{x\sim X}\sum_{l = 1}^{L}\sum_{s = 1}^{S_{l}}l (\hat{z_{l}^{s}}, z_{l}^{s}, z_{l}^{S\setminus s}).
    \label{patchnce_eq}
\end{equation}

Note that in the above equation, L represents a total of L layers of encoder output feature maps to calculate the contrastive loss, different spatial locations in the feature maps of different layers correspond to different image patches in the original image. The deeper the layer, the larger the corresponding patch. $S_{l}$ represents that the feature maps of the current layer have a total of $S_{l}$ spatial locations. $\hat{z_{l}} = H_{l}(G_{enc}^{l}(G(x)))$(where H stands for MLP) indicates that the generated image is passed through the encoder of the generator, and the feature maps output at the l-th layer of the encoder are obtained by MLP for $\hat{z_{l}}$. $z_{l} = H_{l}(G_{enc}^{l}(x))$ indicates that the real image is input to the encoder of the generator, and the feature maps output at the l-th layer of the encoder are obtained by MLP for $z_{l}$. $\hat{z_{l}^{s}}$ is one spatial location of $\hat{z_{l}}$. $\hat{z_{l}^{s}}$ is query, $z_{l}^{s}$ is positive samples, $z_{l}^{S\setminus s}$ is negative samples, which indicates locations other than s in $z_{l}$.

Each spatial location corresponds to the original map is a patch, and each spatial location in the new feature maps obtained by MLP. H is a vector whose dimension is the number of channels. Since query, positive samples, and negative samples are all vectors, we can give the introduction of InfoNCE formula in

\begin{small}
    \begin{equation}
        l(v, v^{+}, v^{-}) = log[\frac{exp(v\cdot v^{+}/\tau )}{exp(v\cdot v^{+}/\tau ) + \sum _{n = 1}^{N}exp(v\cdot v_{n}^{-}/\tau )}].
        \label{contrastiveeq}
    \end{equation}
\end{small}

Before passing vectors into the above formulas, they need to be normalized to prevent space collapse or expansion. Essentially it is actually a cross-entropy loss function, which can close the distance between the query and the positive samples in the feature space, while pushing away the distance between the query and the negative samples in the feature space. It is actually equivalent to an N+1 class classification problem (classifying query with a class corresponding to one positive sample and a class corresponding to N negative samples), as shown in Figure \ref{fig:contrastive}.

\section{Experiment setting}

In this section, we describe the details of Attention-based CUT\_GAN Implementation, including used Datasets, loss function and evaluation metrics.

\subsection{Datasets}
The experiments are based on the Cityscapes \cite{},which contains a diverse set of stereo video sequences recorded in street scenes from 50 different cities, with high quality pixel-level annotations of 5000 frames in addition to a larger set of 20000 weakly annotated frames. and Playing for Data dataset \cite{Playing_for_Data}, which consists of 24966 densely labelled frames split
into 10 parts for convenience. The class labels are compatible with the CamVid and CityScapes datasets, so that it's easy to use while the target is Cityscapes dataset.

\subsection{Training}
To show the effect of the proposed patch-based contrastive loss, we intentionally match the architecture and hyperparameter settings of CycleGAN, except the loss function. This includes the ResNet-based generator\cite{johnson2016resnet-based} with 9 residual blocks, PatchGAN discriminator\cite{ma2019patchgan}, Least Square GAN loss\cite{mao2017lsgan}, batch size of 1, and Adam optimizer\cite{kingma2014adam} with learning rate 0.002.

Our full model Attention-based CUT\_GAN is trained up to 400 epochs, which is the same as CUT\_GAN settings. Our encoder $G_{enc}$ is the first half of the CycleGAN generator\cite{zhu2017unpaired}. In order to calculate our multi-layer, patch-based contrastive loss, we extract features from 5 layers, which are RGB pixels, the first and second downsampling convolution, and the first and the
fifth residual block. The layers we use correspond to receptive fields of sizes
1 $\times$ 1, 9 $\times$ 9, 15 $\times$ 15, 35 $\times$ 35, and 99 $\times$ 99. For each layer's features, we sample 256 locations arrcording to their significance, and apply 2-layer MLP to acquire 256-dim final features. We normalize the vector by its L2 norm.

\subsection{Loss function}
The loss of Attention-based CUT\_GAN comprises three main parts: least square loss $\L_{GAN}$ \cite{mao2017lsgan}, PatchNCE loss $\L_{PatchNCE}(G, H, X)$ from domain X, PatchNCE loss $\L_{PatchNCE}(G, H, Y)$ from domain Y, which are shown in
\begin{small}
    \begin{equation}
    \begin{split}
        \L _{GAN}(G, D, X, Y) + \lambda _{X}\L_{PatchNCE}(G, H, X) + \\ \lambda _{Y}\L_{PatchNCE}(G, H, Y).
        \label{loss_sum}
    \end{split}
    \end{equation}
\end{small}

We choose $\lambda _{X} = 1$ when we jointly train with the identity loss $\lambda _{Y} = 1$, and choose a larger value $\lambda _{X} = 10$ without the identity loss ($\lambda _{Y} = 0$) to compensate for the absence of the regularizer. Our model is relatively simple compared to recent methods that often use 5-10 losses and hyper-parameters.

\subsection{Evaluation Metrics}
Aimed at assessing visual quality and discovered correspondence, we mainly use three metrics to evaluate the generated images: Frechet Inception Distance, Inception Score and Sliced Wasserstein Distance.

\subsubsection{Frechet Inception Distance}
Frechet Inception Distance(FID) is a measure of similarity between two datasets of images, which empirically estimates the distribution of real and generated images in a deep network space and computes the divergence between them. FID was shown to correlate well with human judgement of visual quality and is most often used to evaluate the quality of samples of Generative Adversarial Networks. The smaller the value of FID, the closer the generated image is to the data distribution of the target domain.

\subsubsection{Inception Score}
Inception Score evaluates the quality of GAN-generated images in terms of both clearness and diversity.

\subsubsection{Sliced Wasserstein Distance}
SWD can measure image distribution mismatches or imbalances without additional labels. The idea is to first obtain the representation of the high-dimensional probability distribution in one dimension by linear mapping, and then calculate the wasserstein distance of the one-dimensional representation of the two probability distributions.

\section{Results and Discussion}
In this section, we show the experimental results of four attention modules separately: Triplet Attention, External Attention, Self-attention and Bottleneck Attention. In addition, the baseline GAN such as CycleGAN and CUT\_GAN are also presented and compared.

\subsection{Qualitative Comparision}

\begin{table*}[ht]
\centering
\begin{tabular}{|c|c|c|c|c|c|c|}
\hline
    & CUT\_GAN & CycleGAN      & Self-attention & External\_attention & BAM    & Triplet        \\ \hline
FID & 55.48    & 52.87           & 51.41          & 52.68               & 51.40  & \textbf{50.55} \\ \hline
IS  & 2.42     & 2.44            & \textbf{2.45}  & 2.32                & 2.42   & 2.31           \\ \hline
SWD & 395.78   & \textbf{287.18} & 354.234        & 390.29              & 388.73 & 346.21         \\ \hline
\end{tabular}
\caption{Qualitative Comparision of baselines(CUT\_GAN and CycleGAN) and Attention-based CUT\_GAN with four different attention modules}
\label{tab:results}
\end{table*}

Table \ref{tab:results} shows ablation results for the effect of each attention module in our Attention-based CUT\_GAN. We summarize the Fréchet Inception Distance(FID), Inception Score(IS) and Wasserstein Distance(SWD) results on Attention-based CUT\_GAN with four different attention modules. In addition, CUT\_GAN and CycleGAN are shown as baselines. For the evaluation metric FID(less is better), Triplet Attention module is the best, which obtains 50.55. For the evaluation metric IS(more is better), Self-Attention module is the best, which achieves 2.45. For the evaluation metric SWD(less is better), CycleGAN is the best, which reaches only 287.18. 

\subsection{Discussion}
We use Fréchet Inception Distance(FID) and Sliced Wasserstein Distance (SWD) to evaluate the quality of translated images. FID and SWD both measure the distance between two distributions of real and generated images, and lower indicate the generated image is similar to the real one. In addition, we use Inception Score as evaluation metrics, which considers both image quality and diversity.

According to attention mechanisms, there are four settings in our model, Triplet Attention, External Attention, Self-attention and Bottleneck Attention. For the metric of FID, the translated results of Triplet-Attention based model are more realistic than other methods. For the metric of IS, the translated results of Self-Attention based model performs the best than other methods. However, for the metric of SWD, our proposed Attention-based CUT\_GAN not exceeds CycleGAN. Further, considering the three evaluation metrics together, our Attention-based CUT\_GAN model (no matter which attention mechanism is applied) performs better than the original CUT\_GAN model. Moreover, our method does not add extra model parameters in both generator G and discriminator D.

Compared to baseline methods, our Attention-based CUT\_GAN model has the ability to translate the domain-relevant features accurately. Further, the position and structure of the objects in the generated image are also maintained partially invariant. For example, there are a large number of trees being generated in baseline models, however, this is not present in the simulated image(source domain). This problem is mitigated in our proposed model.

\section{Conclusion}
This paper focuses on solving the third step towards the photo-realistic virtualization of road objects, namely the intelligent post-processing algorithms. Firstly, this paper proposes a Generative adversarial network to realize unpaired image-to-image translation and transformation between the simulated domain and the real domain. Secondly, we propose an Attention Mechanism Module for cross-domain contrastive learning in the task of Image-to-Image Translation. Instead of randomly selecting the anchor, positive and negative samples to compute the contrastive loss, we measure the significance of source domain features and select them based on significance so that the constraint becomes more relevant for the domain translation. Finally, we experimentally prove that our Attention-based CUT\_GAN outperforms the original CUT\_GAN and increases the mutual information between the source domain and target domain.

For the Attention Mechanism Module, we apply four different attention mechanisms: Triplet Attention, External Attention, Self-attention and Bottleneck Attention. All settings outperform the original CUT\_GAN without adding extra model parameters in both generator and discriminator. Besides, we process all generated images and convert them into training datasets of semantic segmentation, in order to verify the practical value of this thesis. After training, the semantic segmentation network can achieve higher accuracy when tested with real images. However, it is still not good as the network is trained with real images. We demonstrated the effectiveness of introducing the attention mechanism module, but at the same time, there still has a lot of work to be done to improve it.

\bibliographystyle{IEEEbib}
\bibliography{icme2023template}

\begin{thebibliography}{10}

\bibitem{goodfellow2020generative}
Ian Goodfellow, Jean Pouget-Abadie, Mehdi Mirza, Bing Xu, David Warde-Farley,
  Sherjil Ozair, Aaron Courville, and Yoshua Bengio,
\newblock ``Generative adversarial networks,''
\newblock {\em Communications of the ACM}, vol. 63, no. 11, pp. 139--144, 2020.

\bibitem{perarnau2016edit}
Guim Perarnau, Joost Van De~Weijer, Bogdan Raducanu, and Jose~M {\'A}lvarez,
\newblock ``Invertible conditional gans for image editing,''
\newblock {\em arXiv preprint arXiv:1611.06355}, 2016.

\bibitem{radford2015dcgan}
Alec Radford, Luke Metz, and Soumith Chintala,
\newblock ``Unsupervised representation learning with deep convolutional
  generative adversarial networks,''
\newblock {\em arXiv preprint arXiv:1511.06434}, 2015.

\bibitem{mao2017lsgan}
Xudong Mao, Qing Li, Haoran Xie, Raymond~YK Lau, Zhen Wang, and Stephen
  Paul~Smolley,
\newblock ``Least squares generative adversarial networks,''
\newblock in {\em Proceedings of the IEEE international conference on computer
  vision}, 2017, pp. 2794--2802.

\bibitem{zhu2017unpaired}
Jun-Yan Zhu, Taesung Park, Phillip Isola, and Alexei~A Efros,
\newblock ``Unpaired image-to-image translation using cycle-consistent
  adversarial networks,''
\newblock in {\em Proceedings of the IEEE international conference on computer
  vision}, 2017, pp. 2223--2232.

\bibitem{brock2018biggan}
Andrew Brock, Jeff Donahue, and Karen Simonyan,
\newblock ``Large scale gan training for high fidelity natural image
  synthesis,''
\newblock {\em arXiv preprint arXiv:1809.11096}, 2018.

\bibitem{isola2017image}
Phillip Isola, Jun-Yan Zhu, Tinghui Zhou, and Alexei~A Efros,
\newblock ``Image-to-image translation with conditional adversarial networks,''
\newblock in {\em IEEE Conference on Computer Vision and Pattern Recognition
  (CVPR)}, 2017.

\bibitem{odena2017conditional}
Augustus Odena, Christopher Olah, and Jonathon Shlens,
\newblock ``Conditional image synthesis with auxiliary classifier gans,''
\newblock in {\em International conference on machine learning}. PMLR, 2017,
  pp. 2642--2651.

\bibitem{liu2021ssl}
Xiao Liu, Fanjin Zhang, Zhenyu Hou, Li~Mian, Zhaoyu Wang, Jing Zhang, and Jie
  Tang,
\newblock ``Self-supervised learning: Generative or contrastive,''
\newblock {\em IEEE Transactions on Knowledge and Data Engineering}, 2021.

\bibitem{kang2020contragan}
Minguk Kang and Jaesik Park,
\newblock ``Contragan: Contrastive learning for conditional image generation,''
\newblock {\em Advances in Neural Information Processing Systems}, vol. 33, pp.
  21357--21369, 2020.

\bibitem{johnson2016resnet-based}
Justin Johnson, Alexandre Alahi, and Li~Fei-Fei,
\newblock ``Perceptual losses for real-time style transfer and
  super-resolution,''
\newblock in {\em European conference on computer vision}. Springer, 2016, pp.
  694--711.

\bibitem{ulyanov2016instance}
Dmitry Ulyanov, Andrea Vedaldi, and Victor Lempitsky,
\newblock ``Instance normalization: The missing ingredient for fast
  stylization,''
\newblock {\em arXiv preprint arXiv:1607.08022}, 2016.

\bibitem{ioffe2015batch}
Sergey Ioffe and Christian Szegedy,
\newblock ``Batch normalization: Accelerating deep network training by reducing
  internal covariate shift,''
\newblock in {\em International conference on machine learning}. PMLR, 2015,
  pp. 448--456.

\bibitem{maas2013relu}
Andrew~L Maas, Awni~Y Hannun, Andrew~Y Ng, et~al.,
\newblock ``Rectifier nonlinearities improve neural network acoustic models,''
\newblock in {\em Proc. icml}. Atlanta, Georgia, USA, 2013, vol.~30, p.~3.

\bibitem{park2020cut}
Taesung Park, Alexei~A. Efros, Richard Zhang, and Jun-Yan Zhu,
\newblock ``Contrastive learning for unpaired image-to-image translation,''
\newblock in {\em European Conference on Computer Vision}, 2020.

\bibitem{ma2019patchgan}
Jinlin Ma, Meng Wei, Ziping Ma, Li~Shi, and Kai Zhu,
\newblock ``Retinal vessel segmentation based on generative adversarial network
  and dilated convolution,''
\newblock in {\em 2019 14th International Conference on Computer Science \&
  Education (ICCSE)}. IEEE, 2019, pp. 282--287.

\bibitem{gutmann2010nce}
Michael Gutmann and Aapo Hyv{\"a}rinen,
\newblock ``Noise-contrastive estimation: A new estimation principle for
  unnormalized statistical models,''
\newblock in {\em Proceedings of the thirteenth international conference on
  artificial intelligence and statistics}. JMLR Workshop and Conference
  Proceedings, 2010, pp. 297--304.

\bibitem{chen2016infogan}
Xi~Chen, Yan Duan, Rein Houthooft, John Schulman, Ilya Sutskever, and Pieter
  Abbeel,
\newblock ``Infogan: Interpretable representation learning by information
  maximizing generative adversarial nets,''
\newblock {\em Advances in neural information processing systems}, vol. 29,
  2016.

\bibitem{Playing_for_Data}
Stephan~R. Richter, Vibhav Vineet, Stefan Roth, and Vladlen Koltun,
\newblock ``Playing for data: {G}round truth from computer games,''
\newblock in {\em European Conference on Computer Vision (ECCV)}, Bastian
  Leibe, Jiri Matas, Nicu Sebe, and Max Welling, Eds. 2016, vol. 9906 of {\em
  LNCS}, pp. 102--118, Springer International Publishing.

\bibitem{kingma2014adam}
P~Kingma~Diederik and Jimmy~Ba Adam,
\newblock ``A method for stochastic optimization,''
\newblock {\em arXiv preprint arXiv:1412.6980}, 2014.

\end{thebibliography}

\end{document}